\documentclass[11pt]{article}

\usepackage[preprint]{acl}

\usepackage{times}
\usepackage{latexsym}

\usepackage[T1]{fontenc}

\usepackage[utf8]{inputenc}

\usepackage[russian,english]{babel}

\usepackage{microtype}

\usepackage{inconsolata}

\usepackage{graphicx}

\usepackage{booktabs}

\title{DHPLT: large-scale multilingual diachronic corpora and word representations for semantic change modelling}

\author{
\textbf{Mariia Fedorova\textsuperscript{1}},
\textbf{Andrey Kutuzov\textsuperscript{1}},
\textbf{Khonzoda Umarova\textsuperscript{2}}\\
\texttt{mariiaf@ifi.uio.no}, \texttt{andreku@ifi.uio.no}, \texttt{ku47@cornell.edu}\\
\textit{(equal contribution, authors sorted in alphabetical order)}
\\
\textsuperscript{1}University of Oslo (Norway),
\textsuperscript{2}Cornell University (USA)
}

\begin{document}
\maketitle
\begin{abstract}
In this resource paper, we present DHPLT, an open collection of diachronic corpora in 41 diverse languages. DHPLT is based on the web-crawled HPLT datasets; we use web crawl timestamps as the approximate signal of document creation time. The collection covers three time periods: 2011-2015, 2020-2021 and 2024-present (1 million documents per time period for each language). We additionally provide pre-computed word type and token embeddings and lexical substitutions for our chosen target words, while at the same time leaving it open for the other researchers to come up with their own target words using the same datasets.

DHPLT aims at filling in the current lack of multilingual diachronic corpora for semantic change modelling (beyond a dozen of high-resource languages). It opens the way for a variety of new experimental setups in this field. 
\end{abstract}

\section{Introduction}\label{sec:intro}

Computational data-driven diachronic semantic change modelling (tracing meaning shifts over time) naturally requires diachronic corpora: that is, texts annotated with their creation date. Once such datasets are obtained, one can compare the usage of target words (or any other linguistic phenomena) in different time periods, using any preferred change modelling method. 

However, diachronic corpora of appropriate size and quality are not easy to find, especially permissively licensed. Most of the current lexical semantic change detection (LSCD) projects operate on the same small set of high-resource languages. For example, the seminal SemEval 2020 Task 1 on LSCD \cite{schlechtweg-etal-2020-semeval} was limited to English, German, Latin and Swedish. Later, LSCD benchmarks based on diachronic corpora for Italian \cite{basile-etal-2019-kronos}, Greek \cite{perrone-etal-2019-gasc}, Russian \cite{kutuzov-pivovarova-2021-three}, Norwegian \cite{kutuzov-etal-2022-nordiachange}, Spanish \cite{d-zamora-reina-etal-2022-black}, Chinese \cite{chen-etal-2023-chiwug-graph}, Japanese \cite{ling-etal-2023-construction}, Finnish \cite{fedorova-etal-2024-axolotl24}, and Slovene \cite{pranjic2024tracking} were presented, but not more than that (not for all languages the corpora themselves are publicly available). The field has to experiment with at most a dozen of languages, with Indo-European family strongly over-represented. This limits the scope of LSCD research, especially on \textit{multilingual} semantic change effects.

To fill in this gap, we release \textbf{DHPLT} (`Diachronic HPLT'): a set of standardized diachronic corpora for 41 languages of 12 different language families. Each language is represented with three time-dependent subsets, containing 1 million documents each. These documents are extracted from the web-crawled datasets by the HPLT project, specifically HPLT v3.0 \cite{oepen2025hplt30largescalemultilingual}: thus, they are basically cleaned and filtered web pages in the target language. We use crawling timestamps as the signal for time period separation (see below).

In addition, we define a set of potentially interesting `target words' for each language. For the DHPLT occurrences of these words, we produce a variety of semantic representations (static \texttt{word2vec} embeddings, token embeddings, lexical substitutions). This allows practitioners to start experimenting with multilingual LSCD immediately, without spending compute on re-creating these representations. At the same time, the availability of the original texts makes it possible to come up with other target word sets. \textbf{All the resources described in this paper are available at \url{https://data.hplt-project.org/three/diachronic/}, sorted by language.}


\section{Diachronic corpora out of HPLT} \label{sec:datasets}
In prior work, diachronic resources for LSCD mostly were produced from existing  historical corpora manually created by linguists: newspaper archives, releases by national libraries, etc. Unfortunately, such resources are nearly non-existent for the majority of world's languages: at least in anything resembling a standardized form. In an ideal world, fragmented national efforts in historical corpora creation could be unified and merged into a multilingual diachronic resource. But the amount of work required for such a project is well beyond the scope of this paper or any research group we are aware of. That's why we instead suggest to rely on the Internet as the source of diachronic data.

World Wide Web contains hundreds of billions of documents in all existing languages and of varying quality (many documents consist of SEO keywords, machine-generated slop or price lists). At least two initiatives are currently crawling the WWW and saving representative slices of its state: Common Crawl\footnote{\url{https://commoncrawl.org/}} (CC) and Internet Archive\footnote{\url{https://archive.org/}} (IA). The HPLT project \cite{burchell-etal-2025-expanded} processes these web crawls by conducting language identification, deduplication, cleaning, etc, to produce language-specific corpora of competitive quality.\footnote{Another project from which one could extract diachronic web corpora is FineWeb 2 \cite{penedo2025fineweb2pipelinescale}.} Importantly, all its datasets are published under the Creative Commons CC0 license. HPLT v3.0\footnote{\url{https://hplt-project.org/datasets/v3.0}} is the specific data release we are using.

HPLT provides a lot of clean documents, but to create a diachronic corpus, we need to know the \textit{date} when the document was created. It is impossible to label all the HPLT documents with the creation date manually. Sometimes, web pages do contain the date of their publication either in plain text or in some structured form. But this data is not reliable: it is perfectly possible for a web document to be published in 2024, but contain text created in 2001. Also, creating parsing rules for all sorts of HTML creation date labels would be an immense effort - with no guarantee that the result will fully reflect the diversity of the Web. Thus, we instead rely on a different time signal: web crawling time stamps. All the HPLT documents can be traced back to specific web crawls and they inherit the `timestamp': that is, the exact date and time when a given web page was downloaded and saved.

Admittedly, these timestamps do not directly map to the creation date of the document: again, it is absolutely possible for CC or IA in 2024 to download a web page created in 2001. But the timestamps do provide an `upper boundary' of the creation date: if some text was crawled in 2015, there is no way for it to have been created later than 2015. Web crawl timestamps allow us to create diachronic datasets of a sort slightly different from `traditional' diachronic corpora. Here, subsets for periods 1, 2 and 3 contain  documents created \textit{no later than 1, 2 or 3} respectively. Importantly, the subset 3 can still contain documents created in the earlier time periods: but not vice versa. For sure, this is less precise than manually labelled historical corpora: but we believe this still can be an important source of diachronic text data.

We 
aim
at a diachronic dataset with more than two time periods, since this makes it possible to conduct research in long-term multipoint dynamics of semantic change \cite{kutuzov-pivovarova-2021-three}. We also would like our time periods to be as comparable as possible in terms of the amount of data, and to be separated by at least some `gaps', since this makes it easier to detect semantic change \cite{giulianelli-etal-2022-fire}. In order to choose the exact temporal spans, we analyse the distribution of documents in the HPLT v3.0 datasets by the year of crawling. Figure~\ref{fig:year_counts} in the Appendix shows these numbers for English and Georgian as an example. Our main observations 
are
that 1) 2011 is the earliest crawl year, and the number of documents remains relatively low until ~2017; 2)  much more documents were crawled in 2020 and after, with peaks in 2020 and 2024 (the latest crawl year).\footnote{Interestingly, 2013 is a rather rare crawl year in HPLT v3.0: many languages have no documents crawled in 2013.}

Based on these observations, we 
come
up with the following three time periods, each 2-4 years long, and with gaps of at least two years: 2011-2015 (`\textit{early time period}'), 2020-2021 (`\textit{Covid time period}'),  2024 (`\textit{most recent crawls}'). The three-time-period structure for the DHPLT is useful for studying and capturing linguistic innovation or the onset of semantic change at different points in time. With the crawl timestamps being the `upper bound' on the document creation time, we can, for instance, observe the rise of the concept of `remote work' in 2020-2021 and then look at its journey in 2024.

Note that our time bins are far from being the only possible choice. We consider them to be a sensible way of temporally splitting the existing HPLT data, but depending on the objective, other splits can make more sense. All the documents in our datasets are accompanied with full timestamps, so anyone can produce their own subsets of DHPLT: for example, more fine-grained. It is also possible to use our open source code\footnote{\url{https://github.com/ltgoslo/scdisc_hplt}} to reproduce DHPLT from the original HPLT data with any desired changes.


We 
produce
three subsets of the HPLT v3.0 datasets containing documents crawled during the time periods above. But first, we 
need
to choose what \textit{languages} DHPLT will contain.

\subsection{Language selection}
The original HPLT v3.0 datasets feature 198 languages, which is way too many for our purposes. Our selection of languages for DHPLT 
is
based on the following criteria:

\begin{enumerate}
    \item Language must have at least 0.5 million documents in each of the time periods above: smaller languages do not provide sufficient amount of data and also are more error-prone with regards to language identification.
    \item There should exist a corresponding HPLT v3.0 \texttt{T5} monolingual encoder-decoder language model\footnote{\url{https://hf.co/collections/HPLT/hplt-30-t5-models}} \cite{oepen2025hplt30largescalemultilingual}: we use these models to generate token embeddings in \ref{sec:representations}.
\end{enumerate}

As a result, we 
come
up with a set of 41 languages. Table~\ref{tab:languages} in the Appendix lists them along with their ISO codes (augmented with the writing system code) and the corresponding number of documents in each of the three time periods.

\subsection{Data extraction pipeline}
For each of the languages, we construct three time-specific corpora by randomly sampling 1 million documents from the HPLT v3.0 dataset corresponding to the given language and time period. Where less than 1 million documents are available, we only sample 0.5 million. The resulting diachronic corpora are published as \texttt{zstd}-compressed JSONL files, following the HPLT format, with the total size $\approx170$ GB (for comparison, the full size of HPLT v3.0 is 50 TB), and $\approx59$ billion words. 

These diachronic corpora can already be used for multilingual LSCD research. However, we also provide more `refined' data for practitioners: namely, semantic representations (which can be computationally expensive for academic researchers to produce) for pre-defined sets of `target words'. They are described in the next sections.  One can think about them as an \textit{example} of what sorts of experimental setups are possible with DHPLT. 

\section{Target word selection}\label{sec:targets}
For each language we select a subset of the vocabulary -- target words -- representations of which would be part of DHPLT. Our primary objective in selecting target words is to narrow down the full corpus vocabulary while keeping as many words that would be of interest to lexical semantic change researchers as possible.

Starting from the T5 model vocabulary corresponding to each given language, we filter out word pieces and infrequent tokens, leaving only words that appear as nouns, verbs or adjectives and are written in the language's main script. Please refer to Appendix~\ref{app:targets} for full details of our target word selection process.

Our selection pipeline yields a set of target words for each DHPLT language, with the average size of $\approx 18 600$ (HPLT T5 models' vocabulary size is $32 768$). ~\autoref{fig:target_word_histograms} shows the distribution of target word counts across languages.

\paragraph{Target lemmas}
We additionally lemmatize each of the resulting target words. For instance, in English distinct target tokens `thread', `Thread', and `threads'  share one common lemma \textbf{`thread'}. We later use lemmas to merge word representations into more linguistically-informed groupings. 



\section{Target word representations}\label{sec:representations}
Once the language-specific target words are defined, we produce a number of different semantic representations for their occurrences in the DHPLT corpora. These representations can be directly used by LSCD practitioners to evaluate or train semantic change models on the three DHPLT time periods.



\subsection{Contextualized word embeddings}
Contextualized token embeddings are widely utilized in lexical semantic-change research~\cite[inter alia]{periti2024systematic, umarova-etal-2025-current}. They can serve both as direct representations that are later averaged into prototypical embeddings~\cite{Periti_2024} and as a basis for constructing clusters corresponding to different `sense nodules'~\cite{martinc2020leveraging, kutuzov2022contextualized}.
For our DHPLT dataset, we obtain encoder embeddings for 1000 randomly sampled occurrences per target word from HPLT v3.0 T5 monolingual models and the XLM-R model \cite{conneau-etal-2020-unsupervised}; we additionally produce encoder embeddings from 100 randomply sampled occurrences per target word from HPLT v3.0 GPT-BERT \cite{charpentier-samuel-2024-bert} monolingual models\footnote{\url{https://hf.co/collections/HPLT/hplt-30-gpt-bert-models}} \cite{oepen2025hplt30largescalemultilingual}.


\subsection{Lexical substitutes}
In addition to language model embeddings, we also consider lexical substitutes as a different kind of contextualized representations. Substitutes-based semantic change quantification methods were shown to do well at both LSCD benchmarks~\cite{card2023substitutionbased, periti-etal-2024-analyzing} and downstream tasks like semantic change discovery~\cite{umarova-etal-2025-current}.

While it is possible to perform masked language modelling with T5 models, the results are not suitable for lexical substitutions generation, since these models were pre-trained with the span masking objective and tend to predict longer sequences rather than single lexemes. Examples can be found in Appendix~\ref{app:t5subst}. For this reason, we use the HPLT v3.0 GPT-BERT models in a way similar to ~\citet{card2023substitutionbased} and~\citet{umarova-etal-2025-current} to represent 100 randomly sampled occurrences of each target word via top-15 substitutes. More details can be found in Appendix~\ref{app:bert}. We also release XLM-R lexical substitutions. The number of target words for which we provide XLM-R embeddings and substitutions is limited by the size  of  the intersections between XLM-R and HPLT T5 tokenizer vocabularies. HPLT v3.0 GPT-BERT models use exactly the same tokenizers as the corresponding HPLT T5 models, so this issue is not relevant for them.

\subsection{Word type embeddings}
Although approaches based on \textit{contextualized token embeddings} (as the ones described above) are the `daily drivers' of modern LSCD researchers \cite{Periti_2024}, we also publish \textit{static type embedding} models trained on the DHPLT corpora. Static word embedding (SWE) models yield one
vector
representation per word type, as opposed to `a representation for each word occurrence' 
from
contextualized models. They were the LSCD mainstream 
until around 2021-2022 \cite{schlechtweg-etal-2020-semeval} and are still often used because of their simplicity and relatively modest compute requirements, both for training and for inference.

We train a SWE model for each language/time period combination using the SGNS architecture, also known as \texttt{word2vec} \cite{word2vec}. For simplicity, we mostly use training hyperparameters from~\citet{aida-bollegala-2025-scdtour}: window size 10, 5 epochs, 5 negative samples. Our embedding size is set to 300, and the model vocabularies are limited to $50 000$ most frequent words.
Before training, the DHPLT documents are filtered to remove punctuation, as well as leading and trailing tabs and whitespaces.

Finally, for each language, the vector spaces of the models trained on time periods $1$ (2011-2015) and $2$ (2020-2021) are \textit{aligned} to the vector space of the model trained on time period $3$ (2024-), so as to make it possible to directly compute similarities between word embeddings in different models. 
We do this
with the standard Procrustes alignment technique \cite{hamilton-etal-2016-diachronic}.

\subsection{Frequency counts}
Finally, we also publish frequency counts of each target word across the three DHPLT time periods. Changes in frequency of word usage coupled with lemma information are some of the very first indicators of changes in the word usage. These counts can also be used to control for frequency effects when quantifying semantic change~\cite{card2023substitutionbased} and for planning compute usage (e.g. when generating lexical substitutions, half of the time is used for finding samples of the 100 least frequent target words, following the Zipf's law \cite{powers-1998-applications}).

\section{Sanity check}\label{sec:cases}
To demonstrate the utility of the DHPLT diachronic corpora, we 
look
at the SWEs of the English word \textit{`AI'} (`artificial intelligence'). Table~\ref{tab:word2vec-ai} shows how the semantics of this term 
drifted.
Back in the beginning of 2010s, it was associated almost exclusively with `AI characters' in video games. A decade later, in 2020-2021, AI starts to be associated with `chatbots' and machine learning, but it is still very much about robots, unmanned vehicles and Internet of Things. Only in the very last period 3 (2024-) we see the much too familiar landscape of LLMs, ChatGPT and `generative AI'. Interestingly, this trajectory is clear even despite the fact that (as noted above), our corpus for period 2 is bound to contain some documents created during the period 1, and the corpus for period 3 surely contains some documents from both 1 and 2.

We observe a similar pattern (Table~\ref{tab:word2vec-ia}) when looking at the SWEs of the Spanish equivalent of the word: `IA' (`inteligencia artificial'). In 2010s, the word often appears in the context of gaming: `jugabilidad', `PS', `BETA', etc. Then in 2020-2021, we start seeing words like `algoritmos' and `tecnología(s)' among closest semantic neighbours. These words follow more semantically-similar English-words such as `AI', `artificial', `learning', etc, which are likely to still appear in the Spanish DHPLT corpora as part of names and titles. Finally, in 2024 `IA' starts being associated with `generativa' and `ChatGPT'.  Very similar trends are found in our SWEs trained on Russian DHPLT documents (Table~\ref{tab:word2vec-rus}).

For T5 encoder embeddings, we calculate average pairwise distance between different time period representations \cite{kutuzov2022contextualized} for the English lemmas `ai', `remote', `legislative', and `jurisdiction'. The exact scores are to be found in the Appendix \ref{app:cases}. The change of `ai' semantics is the largest and corresponds to the aforementioned SWE findings, while the changes of `legislative', and `jurisdiction', which are terms from the conservative legal domain, are the smallest. The degree of `remote' change is somewhere in between and is the largest between 2011-2015 and 2020-2021 (`Covid') periods, when this word began to refer specifically to remote work rather than other contexts\footnote{\url{https://languages.oup.com/word-of-the-year/2020/}}. These observations hold for Spanish DHPLT corpora as well (Table~\ref{tab:apd-esp}).

\section{Conclusion}\label{sec:conclusion}
\label{sec:bibtex}

We 
present
DHPLT (`Diachronic HPLT'): an open collection of large-scale diachronic corpora in 41 languages of 12 different language families. It is based on the web-crawled HPLT v3.0 datasets \cite{oepen2025hplt30largescalemultilingual}, using web crawl timestamp as the temporal signal. The collection covers three time periods: 2011-2015, 2020-2021 and 2024.
We augment DHPLT with pre-computed token-level semantic representations for language-specific sets of target words, to make it easier for practitioners to start experimenting with our corpora. Finally, we provide aligned static (type-based) word embedding models for each language and time period.

DHPLT (partially) addresses the lack of multilingual diachronic corpora in the LSCD field. We hope it will help making the landscape of historical language change modelling more rich and diverse. It should be especially relevant for studies in semantic change discovery. 

\section*{Limitations}

The main limitation of DHPLT is the source of temporal signal: that is, web crawl timestamps. As described above, timestamp $X$ on a document does not guarantee that the text in this document was not created in the time periods $<X$ (earlier than $X$). It guarantees only that the text was not created in the time periods $>X$ (later than $X$). This difference compared to traditional diachronic corpora should be kept in mind when working with DHPLT.

Another limitation is that we provide only some of the possible types of semantic representations, and only for selected sets of target word, not for \textit{all} the words in each of our languages. This is inevitable, given compute and storage space constraints. DHPLT allows practitioners to come up with their own sets of target words, or even conducts semantic change discovery experiments on the entire corpus. Our representations were obtained from models that do not incorporate any temporality, thus, e.g. masked language modeling predictions for 2011-2015 can include proper names that in fact first emerged in later periods. For future work, one might employ approach from \cite{fittschen2025pretraininglanguagemodelsdiachronic} and pre-train models on texts from corresponding periods only.

Finally, in this paper we only introduce the DHPLT dataset; we leave conducting full-scale semantic change discovery on it for future work.

\section*{Acknowledgments}

The computations were performed on resources provided by Sigma2 - the National Infrastructure for High-Performance Computing and Data Storage in Norway.
This work was also in part supported by a gift from Google. Any opinions, findings, and conclusions or recommendations
expressed in this material are those of the authors and do not
necessarily reflect the views of Google.


\bibliography{anthology,custom}

\appendix

\section{DHPLT datasets}
\label{app:data}

\begin{figure}[h]
    \centering
    \includegraphics[width=0.45\linewidth]{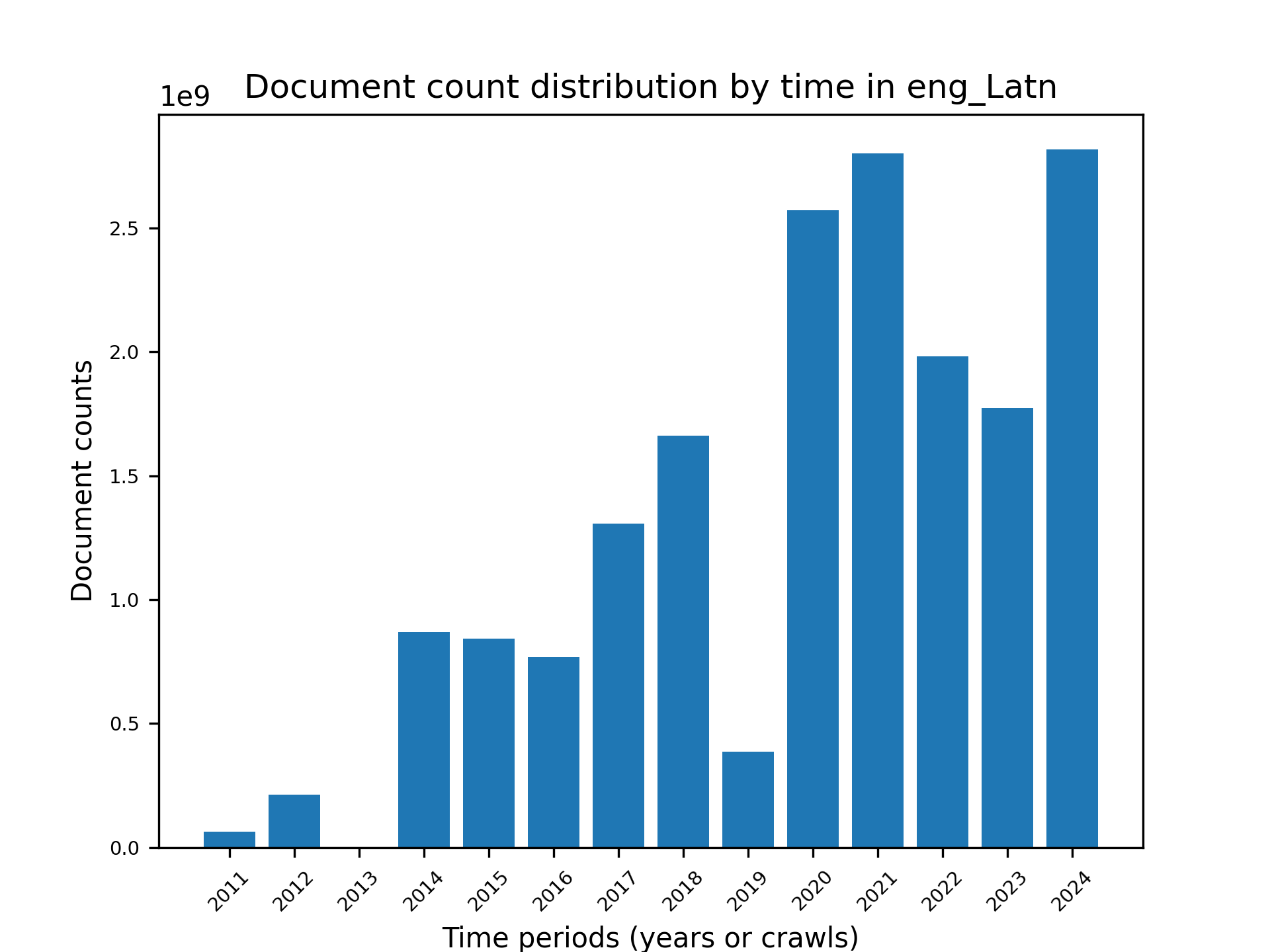}
     \includegraphics[width=0.45\linewidth]{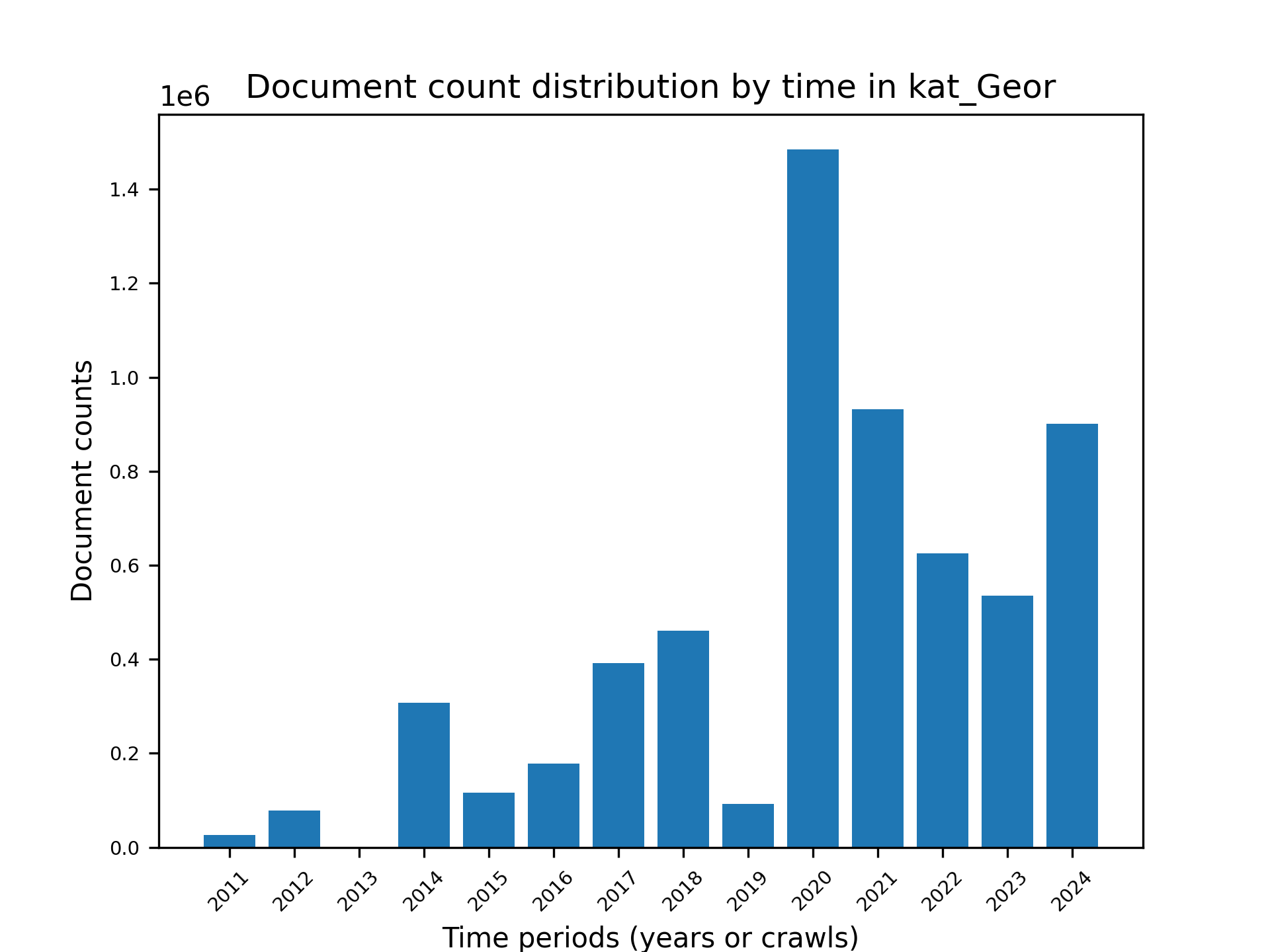}
    \caption{Number of documents per crawl year in the HPLT v3.0 datasets: English (left) and Georgian (right).}
    \label{fig:year_counts}
\end{figure}

Figure~\ref{fig:year_counts} shows the number of documents from different crawls for English and Georgian languages in HPLT v3.0. 

DHPLT files contain one document per line, with the following data fields:

\begin{itemize}
    \item \texttt{id}: unique document identifier, can be used to link back to the original HPLT dataset,
    \item \texttt{ts}: timestamp, the exact date and time when the document was crawled from the Web,
    \item \texttt{text}: the actual document body, split into `segments' (in most cases equal to paragraphs) with line break symbols,
    \item \texttt{doc\_scores}: a list of integer `quality scores' assigned to each segment of the document;  to produce these scores, HPLT project employs heuristics from \texttt{Web Docs Scorer} (WDS).\footnote{\url{https://github.com/pablop16n/web-docs-scorer}}
\end{itemize}

Table~\ref{tab:languages} lists all the DHPLT languages and their statistics.

\begin{table}[h]
    \centering
\resizebox{\linewidth}{!}{
    \begin{tabular}{lll|ccc}
\toprule
\textbf{Language} & \textbf{ISO Code} & \textbf{Family} & \textbf{1} & \textbf{2} & \textbf{3} \\
\midrule
Albanian & als\_Latn & Indo-European  & 0.5M & 1M & 1M \\
Arabic & arb\_Arab & Afro-Asiatic  & 1M & 1M & 1M \\
Bosnian & bos\_Latn & Indo-European  & 1M & 1M & 1M \\
Bulgarian & bul\_Cyrl & Indo-European   & 1M & 1M & 1M \\
Catalan & cat\_Latn & Indo-European   & 1M & 1M & 1M \\
Czech & ces\_Latn & Indo-European  & 1M & 1M & 1M \\
Chinese & cmn\_Hans & Sino-Tibetan & 1M & 1M & 1M \\
Danish & dan\_Latn & Indo-European  & 1M & 1M & 1M \\
German & deu\_Latn & Indo-European  & 1M & 1M & 1M \\
Estonian & ekk\_Latn & Uralic & 0.5M & 1M & 1M \\
Greek & ell\_Grek & Indo-European   & 1M & 1M & 1M \\
English & eng\_Latn & Indo-European  & 1M & 1M & 1M \\
Finnish & fin\_Latn & Uralic  & 1M & 1M & 1M \\
French & fra\_Latn & Indo-European  & 1M & 1M & 1M \\
Hebrew & heb\_Hebr & Afro-Asiatic & 1M & 1M & 1M \\
Croatian & hrv\_Latn & Indo-European  & 1M & 1M & 1M \\
Hungarian & hun\_Latn & Uralic & 1M & 1M & 1M \\
Armenian & hye\_Armn & Indo-European & 0.5M & 1M & 0.5M \\
Indonesian & ind\_Latn & Austronesian  & 1M & 1M & 1M \\
Italian & ita\_Latn & Indo-European  & 1M & 1M & 1M \\
Japanese & jpn\_Jpan & Japanese  & 1M & 1M & 1M \\
Georgian & kat\_Geor & Kartvelian  & 0.5M & 1M & 0.5M \\
Korean & kor\_Hang & Korean  & 1M & 1M & 1M \\
Lithuanian & lit\_Latn & Indo-European  & 1M & 1M & 1M \\
Latvian & lvs\_Latn & Indo-European  & 0.5M & 1M & 1M \\
Macedonian & mkd\_Cyrl & Indo-European  & 0.5M & 1M & 1M \\
Dutch & nld\_Latn & Indo-European  & 1M & 1M & 1M \\
Norwegian & nob\_Latn & Indo-European  & 1M & 1M & 1M \\
Polish & pol\_Latn & Indo-European  & 1M & 1M & 1M \\
Portuguese & por\_Latn & Indo-European  & 1M & 1M & 1M \\
Romanian & ron\_Latn & Indo-European  & 1M & 1M & 1M \\
Russian & rus\_Cyrl & Indo-European  & 1M & 1M & 1M \\
Slovak & slk\_Latn & Indo-European  & 1M & 1M & 1M \\
Slovenian & slv\_Latn & Indo-European  & 1M & 1M & 1M \\
Spanish & spa\_Latn & Indo-European  & 1M & 1M & 1M \\
Swedish & swe\_Latn & Indo-European  & 1M & 1M & 1M \\
Tamil & tam\_Taml & Dravidian  & 0.5M & 1M & 1M \\
Thai & tha\_Thai & Tai-Kadai  & 1M & 1M & 1M \\
Turkish & tur\_Latn & Altaic  & 1M & 1M & 1M \\
Ukrainian & ukr\_Cyrl & Indo-European  & 1M & 1M & 1M \\
Vietnamese & vie\_Latn & Austro-Asiatic  & 1M & 1M & 1M \\
\bottomrule
    \end{tabular}
}
    \caption{DHPLT languages, writing systems, language families and historical period sizes (in millions of documents).}
    \label{tab:languages}
\end{table}

\section{DHPLT target words}
\label{app:targets}
Below we provide technical details on selection of the target words.

For a given language $L$, we start from the vocabulary $V_{T5_L}$ of the corresponding T5 model from the HPLT v3 T5 model collection. These models were pre-trained on documents in specific languages from the original HPLT v3.0 dataset. We assume that words which don't appear as their own token in the corresponding T5 vocabulary are not frequent enough, so we omit them. 

Further, we exclude tokens from $V_{T5_L}$ that are word pieces or non-words. To do this, we first employ a heuristic for identifying full words. For most languages after removing punctuation we split the documents from diachronic corpora by whitespace, and count occurrences of such `full words'. For languages where splitting by whitespace doesn't make sense, we use specific splitters. For Japanese,  we employ \texttt{fugashi}\footnote{\url{https://github.com/polm/fugashi}} library that utilizes dictionaries for tokenization. For simplified Chinese,  we segment words in text using \texttt{jieba}\footnote{\url{https://github.com/fxsjy/jieba}} library. Finally, for Thai, we go with word tokenization via \texttt{pythainlp}\footnote{\url{https://github.com/PyThaiNLP/pythainlp}} library. 

Next, we count occurrences of such `full words' across the diachronic corpora and filter out infrequent terms using a minimum frequency threshold. Thus, we only keep a token from $V_{T5_L}$ if it appears as a `full word' at least 10 times in each of the three time periods (i.e., at least 30 times across all diachronic corpora in that language). Note that we ignore case when counting frequencies: e.g., if `operation' and `OPERATION' each occurs 5 times in the corpus, we keep both.

Further, following~\citet{kurtyigit-etal-2021-lexical}, we also limit target words to only nouns, verbs, or adjectives. We use Stanza \cite{qi-etal-2020-stanza} part-of-speech taggers for all languages except Macedonian, for which we use \texttt{classla} \cite{ljubesic_2024_13936406}.
In cases where the tagger identifies a token as a proper noun but its lower-cased version is tagged as a noun, the `NOUN' tag takes precedence. 
Additionally, with the exception of Chinese, Japanese and Korean, we remove single character tokens in all languages.

Finally, we ensure that words in the target set are written in this language's main script. Following the approach similar to \texttt{alphabet-detector}\footnote{\url{https://github.com/EliFinkelshteyn/alphabet-detector}}, we use the Unicode names of characters in a target word candidate to verify its script. For instance, for English, we check that each character in each word from the target set is from Latin Unicode blocks, while for Japanese, we check that characters are either Hiragana, Katakana, or Kanji. We exclude words in which at least one character doesn't belong to the expected script.

\begin{figure}[h]
    \centering
    \includegraphics[width=\linewidth]{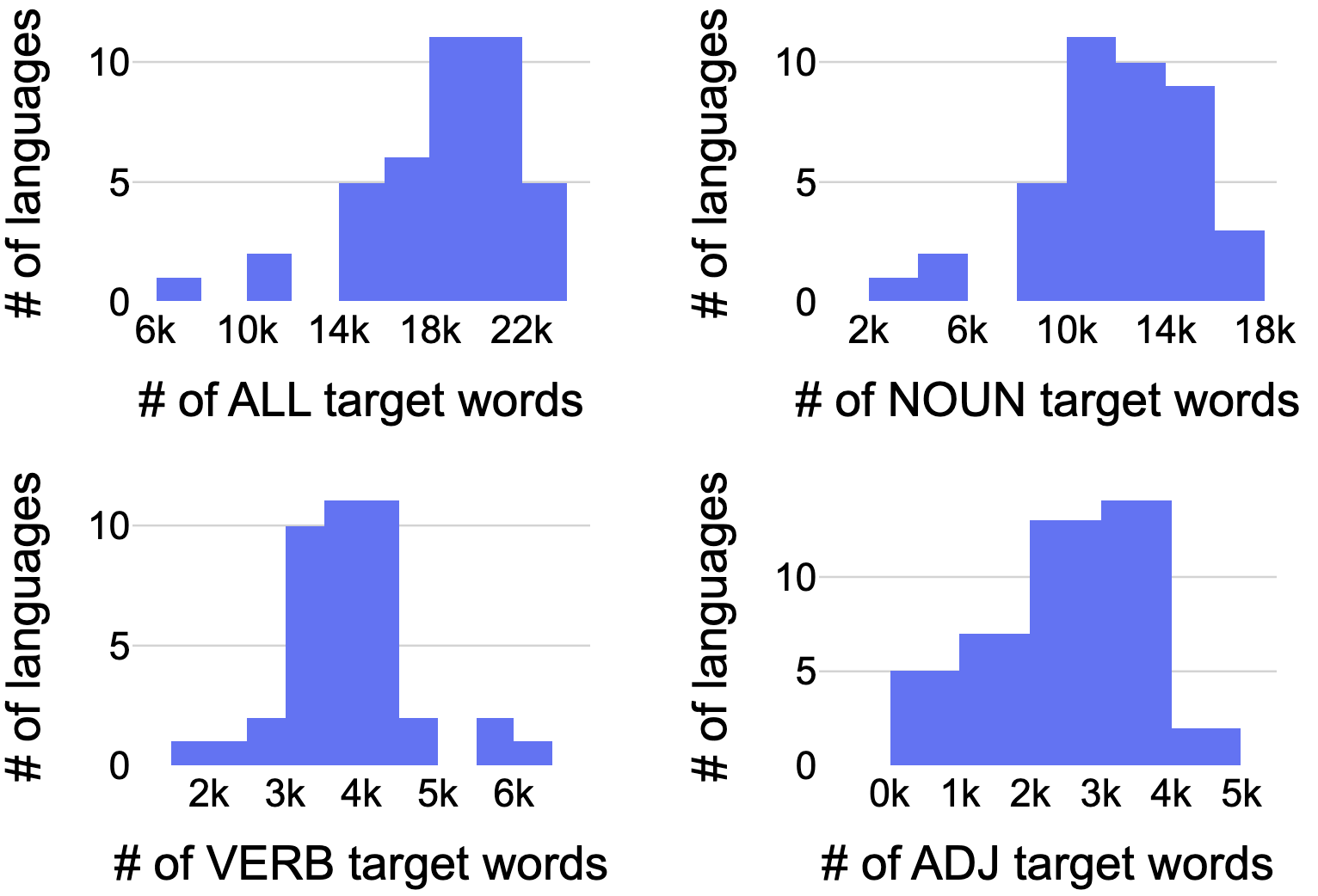}
    \caption{Number of target words across 41 languages for all target words (top left), target words that are nouns (top right), verbs (bottom left), and adjectives (bottom right).}
    \label{fig:target_word_histograms}
\end{figure}

Figure~\ref{fig:target_word_histograms} shows the distribution of target word counts across languages in DHPLT.

\section{T5 substitutions}
\label{app:t5subst}

We use the same input format in our experiments, as shown in T5 model cards\footnote{\url{https://huggingface.co/collections/HPLT/hplt-30-t5-models}}.
The reasons for not using T5 to generate lexical substitutions are :

\begin{itemize}
    \item both encoder and decoder parts are required, duplicating the compute usage
    \item while it is technically possible to obtain not only the one most probable, but also top-k predictions with \texttt{beam search}~\cite{freitag-al-onaizan-2017-beam}, in practice the difference between predictions is too vague to make them useful. For example, the input \textit{I remember we  [MASK\_1] it but still - do we use the decoder part because the final hidden states of the encoder part are not mapped to vocabulary logits?} (with the word `discussed' behind the mask) yields the predictions `used the decoder part to do', `used the decoder part for', and `used the decoder part before'. Such small differences in  generated representations are rather a problem for LSCD than an advantage, as shown in \citet{fedorova-etal-2024-definition}.
    \item predictions do not necessarily represent the semantics of the target word and tend to repeat other terms from the sentence, which is also observed on the aforementioned example
    \item longer inputs from real-world HPLT documents yield even longer predictions, for example, 
    \textit{How do you know the apples you are using for hard cider are ripe? Maybe, you would [MASK\_1] me to define ripe. Is ripe defined by the ideal time to harvest an apple, to eat an apple, or to press an apple. We could even consider the question of ripeness for cooking apples. In my ... Continue reading When are apples ripe?} yields \textit{`say that the apples you are using for hard cider are ripe. But I don't think that is the right way for'}
\end{itemize}

\section{BERT representations}
\label{app:bert}
HPLT 3.0 T5s and HPLT 3.0 GPT-BERTs use the same tokenizer vocabulary for each language, so each target word has representations produced by both models.

\section{Examples}
\label{app:cases}

\subsection{Sanity check for T5 embeddings}
\label{app: t5}
Tables~\ref{tab:apd} and~\ref{tab:apd-esp} show change degrees of the English (`ai', `remote', `legislative', `jurisdiction') and Spanish (`ia', `remoto', `legislativo', `jurisdicción') words, according to the average pairwise distance, APD method \cite{fedorova-etal-2024-definition}, on their respective T5 token embeddings.

\begin{table}[h]
    \centering
    \resizebox{\columnwidth}{!}{%
    \begin{tabular}{l|llll}
    \toprule
        \textbf{Period pairs} & \textbf{`ai'} & \textbf{`remote'} & \textbf{`legislative'} & 
        \textbf{`jurisdiction'} \\
    \midrule
        1 to 2 & 0.5533 & 0.4586 & 0.4117 & 0.4495 \\
        1 to 3 & \textbf{0.5646} & 0.4619 & 0.4141 & 0.4497 \\
        2 to 3 & 0.48 & 0.4548 & 0.4191 & 0.4351 \\
    \bottomrule
    \end{tabular}%
    }
    \caption{Average pairwise distances for several English target words calculated on T5 encoder embeddings.}
    \label{tab:apd}
\end{table}

\begin{table}[h]
    \centering
    \resizebox{\columnwidth}{!}{%
    \begin{tabular}{l|llll}
    \toprule
        \textbf{Period pairs} & \textbf{`ia'} & \textbf{`remoto'} & \textbf{`legislativo'} & 
        \textbf{`jurisdicción'} \\
    \midrule
        1 to 2 & 0.5733 & 0.5104 & 0.4031 & 0.4470 \\
        1 to 3 & 0.5763 & 0.4955 & 0.3925 & 0.4438 \\
        2 to 3 & \textbf{0.5810} & 0.4821 & 0.3979 & 0.4423 \\
    \bottomrule
    \end{tabular}%
    }
    \caption{Average pairwise distances for several Spanish target words calculated on T5 encoder embeddings.}
    \label{tab:apd-esp}
\end{table}

\subsection{Sanity check of HPLT 3.0 GPT-BERT substitutions}
\label{app: bert}

We perform manual analysis of the same 4 English words as in Section~\ref{sec:cases}. The observations obtained with SWE models still hold: in 2011-2015, the words predicted as substitutions for `ai' are either non-technical, or related to games or cars. In 2020-2021, a wider range of technologies is mentioned, including `IoT', `NLP', `robotics', `animation', etc. Also a lot of terms reflecting the social influence of AI emerge: `cybersecurity', `humanity', `innovation', names of states and companies. Finally, in 2024, the trend of discussing social consequences continues: `elite', `censorship', `communism', `scammers', `capitalism'; much less technical terms and much more human-related ones are observed. There are also mentions of spheres which traditionally were human-dominated but has become automated recently: `art', `healthcare' etc. Surprisingly, we don't observe many `LLM'-related terms among GPT-BERT's predictions, but rather a shift from the optimistic perception of AI to the pessimistic one.

In 2011-2015, `remote' is associated with networks and being spatially (geographically) distant. In 2020-2021, `virtual' frequently occurs. In 2024, the associations show a techno-optimistic pattern similar to that of `AI' in 2020-2021: positive job-related adjectives (`skilled', `flexible', `professional'), wider range of technologies and spheres (`satellite' and `healthcare' emerge). We also see terms related to society: state names, `climate', `rural'. 

Substitutions of `jurisdiction' and `legislative' bring no surprises, being related to law throughout all three time periods.

To conclude, representations obtained from contextualized models are sensitive to particular contexts at prediction, and thus capture more fine-grained semantic nuances than SWE models.

\subsection{Sanity check for SWEs}
\label{app: word2vec}

Table~\ref{tab:word2vec-ai} shows the semantic trajectory of the English word `AI' in the DHPLT time periods, according to our static word embedding models (SWEs). Similarly, Table~\ref{tab:word2vec-ia} shows the trajectory for `IA' that stands for `inteligencia artificial' in Spanish across the three DHPLT time periods. Table~\ref{tab:word2vec-rus} does the same for the Russian abbreviation `\foreignlanguage{russian}{ИИ}' (`AI') (the model trained on the first time period does have this word in its vocabulary). 

\begin{table}[h]
    \centering
    \begin{tabular}{l|l|l}
    \toprule
    \textbf{1: 2011-2015} & \textbf{2: 2020-2021} & \textbf{3: 2024-} \\
         \midrule
    multiplayer     & chatbots  & generative \\
    NPCs    & IoT  & AI’s \\
    RPG     & robotics  & GenAI \\
    animations    & RPA  & ChatGPT \\
    FPS     & intelligence  & LLMs \\
    \bottomrule
    \end{tabular}
    \caption{Top 5 nearest neighbours (by cosine similarity) of the English term `AI' in DHPLT static word embedding models by time periods. Case is ignored.}
    \label{tab:word2vec-ai}
\end{table}

\begin{table}[h]
    \begin{tabular}{l|l|l}
    \toprule
    \textbf{1: 2011-2015} & \textbf{2: 2020-2021} & \textbf{3: 2024-} \\
         \midrule
    BETA & AI & generativa\\
    PS & artificial & artificial\\
    AI & algoritmos & AI\\
    jugabilidad & learning & inteligencia\\
    artificial & inteligencia & ChatGPT\\
    \bottomrule
    \end{tabular}
    \caption{Top 5 nearest neighbours (by cosine similarity) of the Spanish term `IA' in DHPLT static word embedding models by time periods. Case is ignored.}
    \label{tab:word2vec-ia}
\end{table}

\begin{table}[h]
    \resizebox{\columnwidth}{!}{%
    \begin{tabular}{l|l}
    \toprule
\textbf{2: 2020-2021} & \textbf{3: 2024-} \\
         \midrule
\foreignlanguage{russian}{интеллект} (intellect)  & \foreignlanguage{russian}{интеллект} (intellect) \\
AI  & \foreignlanguage{russian}{нейросети} (neural networks) \\
\foreignlanguage{russian}{роботов} (robots)  & ChatGPT \\
 \foreignlanguage{russian}{блокчейн} (blockchain)  & AI \\
\foreignlanguage{russian}{алгоритмы} (algorithms)  & \foreignlanguage{russian}{искусственный} (artificial) \\
    \bottomrule
    \end{tabular}%
    }
    \caption{Top 5 nearest neighbours (by cosine similarity) of the Russian term \foreignlanguage{russian}{`ИИ'} (`AI') in DHPLT static word embedding models by time periods. Case is ignored.} The 2011-2015 model does not have the word in its vocabulary (because of low frequency in this time period).
    \label{tab:word2vec-rus}
\end{table}

\end{document}